\documentclass[10pt,twocolumn,letterpaper]{article}

\usepackage{iccv}
\usepackage{times}
\usepackage{epsfig}
\usepackage{graphicx}
\usepackage{amsmath}
\usepackage{amssymb}
\usepackage{caption}
\usepackage{subfigure}

\usepackage{pgfplots}
\usepackage{pgfplotstable}
\pgfplotsset{compat=newest}

\usepackage[pagebackref=true,breaklinks=true,letterpaper=true,colorlinks,bookmarks=false]{hyperref}

\iccvfinalcopy 

\def\iccvPaperID{1839} 

\ificcvfinal\fi
\begin{document}

\title{Learning Visual Clothing Style with Heterogeneous Dyadic Co-occurrences}

\author{
Andreas Veit\thanks{These two authors contributed equally; the order is picked at random.}\textsuperscript{ \ \ 1}, 
Balazs Kovacs\footnotemark[1]\textsuperscript{ \ \  1}, 
Sean Bell\textsuperscript{1}, 
Julian McAuley\textsuperscript{3}, 
Kavita Bala\textsuperscript{1}, 
Serge Belongie\textsuperscript{1,2}\\
\\
\textsuperscript{1} Department of Computer Science, Cornell University \quad \textsuperscript{2} Cornell Tech\\
\textsuperscript{3} Department of Computer Science and Engineering, UC San Diego\\
}

\maketitle

\begin{abstract}
With the rapid proliferation of smart mobile devices, users now take millions of photos every day. These include large numbers of clothing and accessory images. We would like to answer questions like `What outfit goes well with this pair of shoes?' To answer these types of questions, one has to go beyond learning visual similarity and learn a visual notion of compatibility across categories. In this paper, we propose a novel learning framework to help answer these types of questions. The main idea of this framework is to learn a feature transformation from images of items into a latent space that expresses compatibility. For the feature transformation, we use a Siamese Convolutional Neural Network (CNN) architecture, where training examples are pairs of items that are either compatible or incompatible. We model compatibility based on co-occurrence in large-scale user behavior data; in particular co-purchase data from Amazon.com. To learn cross-category fit, we introduce a strategic method to sample training data, where pairs of items are \textit{heterogeneous dyads}, i.e., the two elements of a pair belong to different high-level categories. While this approach is applicable to a wide variety of settings, we focus on the representative problem of learning compatible clothing style. Our results indicate that the proposed framework is capable of learning semantic information about visual style and is able to generate outfits of clothes, with items from different categories, that go well together.
\end{abstract}

\begin{figure}[t]
\begin{center}
   \includegraphics[width=0.85\linewidth]{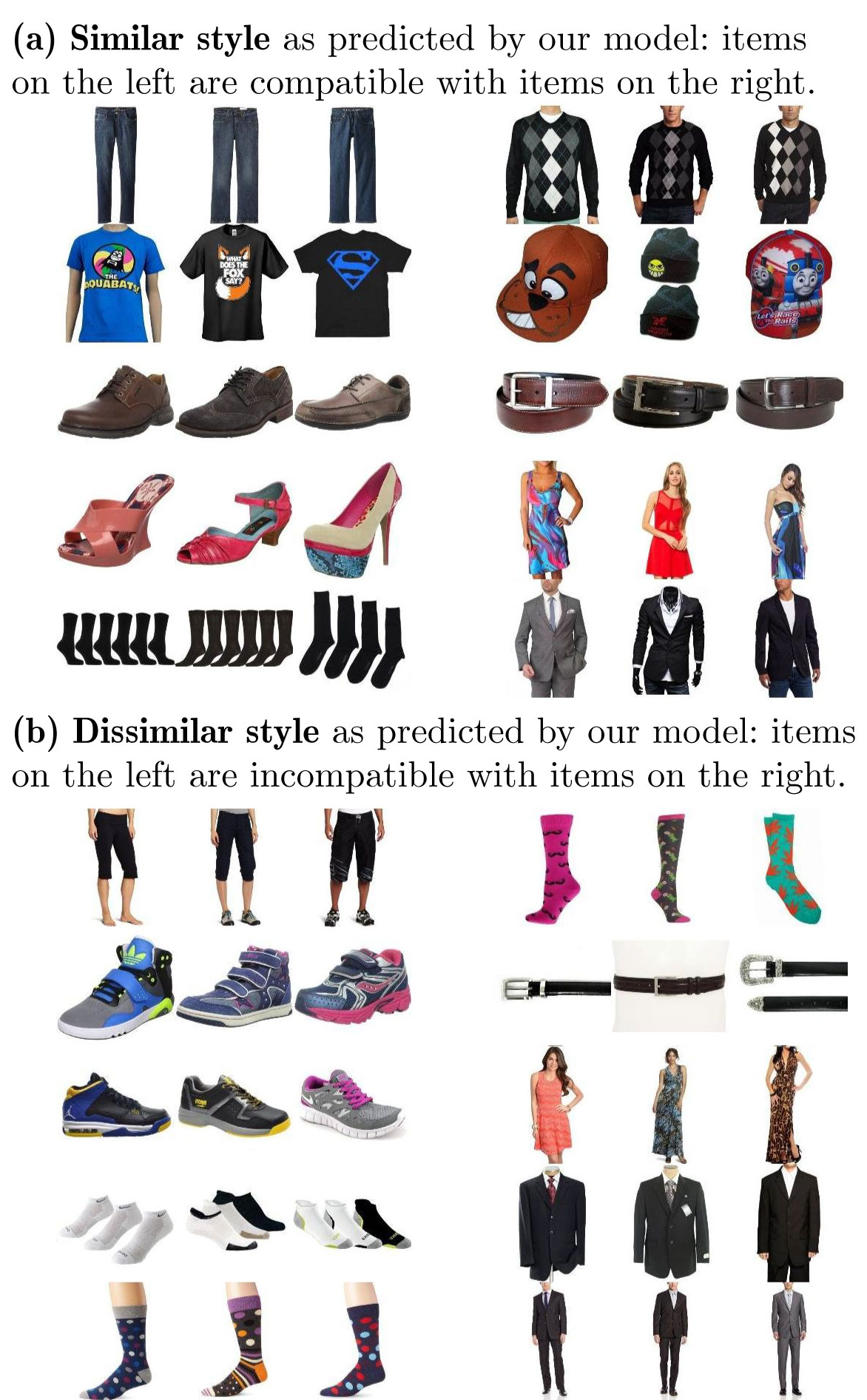}
\end{center}
\vspace{-10pt}
\caption{Example similar and dissimilar items predicted by our model. Each row shows a pair of clusters; items on the same side belong to the same clothing category and cluster. \textbf{(a)}: each row shows two clusters that are stylistically compatible; \textbf{(b)}: each row shows incompatible clusters.}
\label{fig:stylenotion}
\end{figure}

\section{Introduction}
\begin{figure*}[t]
\begin{center}
   \includegraphics[width=1\linewidth]{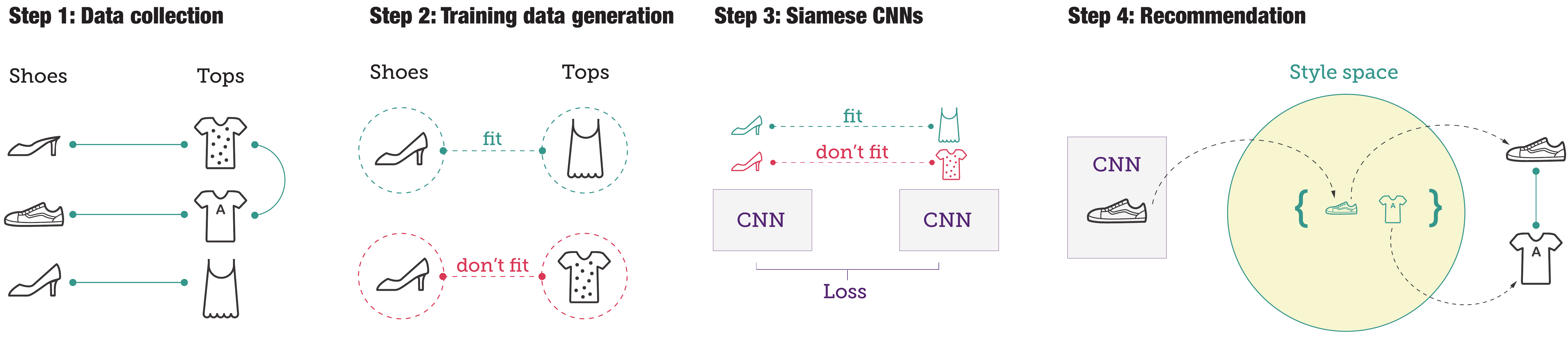}
\end{center}
\vspace{-10pt}
\caption{The proposed framework consists of four key components: \textbf{(1)} The input data comprises item images, category labels and links between items, describing co-occurrences. \textbf{(2)} From the input data, we strategically sample training pairs of items that belong to different categories. \textbf{(3)} We use Siamese CNNs to learn a feature transformation from the image space to the style space. \textbf{(4)} Finally, we use a robust nearest neighbor retrieval to generate outfits of compatible items.}
\label{fig:overview}
\end{figure*}
Smart mobile devices have become an important part of our lives and people use them to take and upload millions of photos every day. Among these photos we can find large numbers of clothing and food images. Naturally, we would like to answer questions like 	``What outfit matches this pair of shoes?'' or ``What desserts would go well along this entr\'{e}e?'' A straightforward approach to answer this type of questions would be to use fine grained recognition of subcategories and attributes, e.g., ``slim dark formal pants,'' with a graph that informs which subcategories match together. 
However, these approaches require significant domain knowledge and do not generalize well to the introduction of new subcategories. Further, they require large datasets with fine grained category labels, which are difficult to collect. Getting domain knowledge and collecting large datasets becomes especially hard in domains like clothing, where fashion collections change every season.

In this paper, we propose a novel learning framework to overcome these challenges and help answer the raised questions.  Our framework allows learning a feature transformation from the images of the items to a latent space, which we call \textit{style space}, so that images of items from different categories that match together are close in the \textit{style space} and items that don't match are far apart. 
Our proposed framework is capable of retrieving bundles of compatible objects. A bundle refers to a set of items from different categories, like shirts, shoes and pants. 
The challenge of this problem is that the bundle of objects come from visually distinct categories. For example, clothing items with completely different visual cues may be similar in our style space, \eg white shirts and black pants. However, this high contrast does not generally imply a stylistic match; for example, white socks tend to not match to black pants. Figure~\ref{fig:stylenotion} shows pairs of items that are very close in the \textit{style space} (top rows) and also pairs that are very far apart (bottom rows). 

The proposed framework consists of four parts. Figure~\ref{fig:overview} provides an illustration of the basic flow. First, the input data comprises item images, category labels and links between items, describing co-occurrences. Then, to learn style across categories, we strategically sample training examples from the input data such that pairs of items are co-occurring \textit{heterogeneous dyads}, i.e., the two items belong to different categories and frequently co-occur. Subsequently, we use Siamese CNNs~\cite{hadsell2006dimensionality} to learn a feature transformation from the image space to the latent style space. Finally, we generate structured bundles of compatible items by querying the learned latent space and retrieving the nearest neighbors from each category to the query item.

To evaluate our learning framework, we use a large-scale dataset from Amazon.com, which was collected by \cite{julian}. As a measure of compatibility between products, we use co-purchase data from Amazon customers. 
In our experiments, we observe that the learned style space indeed expresses extensive semantic information about visual clothing style.
Further, we find that the feature transformation learned with our framework quantitatively outperforms the vanilla ImageNet features \cite{szegedy2014going} as well as the common approach where Siamese CNNs are trained without the proposed strategic sampling of training examples \cite{bell15siggraph, cvpr2015geolocalization}. 

Our main contributions are the following:
\begin{enumerate}
\item 
We propose a new learning framework that combines Siamese CNNs with co-occurrence information as well as category labels. 
\item
We propose a strategic sampling approach for pairwise training data that allows learning compatibility across categories.
\item 
We present a robust nearest neighbor retrieval method for datasets with strong label noise. 
\item
We conduct a user study to understand how users think about style and compatibility. Further, we compare our learning framework against baselines. 
\end{enumerate}


\section{Related work}
Our work is related to different streams of research. We focus this discussion on metric learning and attributes, convolutional neural networks for learning distance metrics and image retrieval as well as learning clothing style. 

\textbf{Metric learning and attributes.}
Metric learning is used to learn a \emph{continuous} high dimensional embedding space. This research field is wide and we refer to the work of Kulis \cite{kulis2012metric} for a comprehensive survey. A different approach is the use of \emph{attributes} that assign semantic labels to specific dimensions or regions in the feature space. An example is Whittle search that uses relative attributes to guide product search \cite{kovashka2012whittlesearch}. 
In contrast with these works, we want to learn a feature transformation from the input image to a similarity metric that does not rely on discrete and pre-defined attributes. 

\textbf{Convolutional neural networks for learning distance metrics and image retrieval.}
Although convolutional neural networks (CNNs) were introduced many years ago \cite{lecun1998gradient}, they have experienced a strong surge in interest in recent years since the success of of Krizhevsky \etal \cite{krizhevsky2012imagenet} in the ILSVRC2012 image classification challenge~\cite{ilsvrc}. We use two of the most successful network architectures, i.e, AlexNet \cite{krizhevsky2012imagenet} and GoogLeNet \cite{szegedy2014going}. Razavian \etal \cite{razavian2014cnn} show that CNNs trained for object classification produce features that can even be used successfully for image instance retrieval. To compare our framework to this approach, we include the vanilla ImageNet GoogLeNet as a baseline in our evaluations. 

Since the introduction of the Siamese setup \cite{hadsell2006dimensionality}, CNNs are increasingly used for metric learning and image retrieval. The advantage of the Siamese setup is that it allows to directly learn a feature transformation from the image space to a latent space of metric distances. This approach has been successfully applied to learn correspondences between images that depict houses from different viewpoints, i.e., street view vs. aerial view, for image geo-localization \cite{cvpr2015geolocalization}. Further, Chopra \etal \cite{chopra2005learning}, Hu \etal \cite{hu2014discriminative} apply Siamese networks in context of face verification.

In this stream of research, the closest work to ours is the work by Bell and Bala \cite{bell15siggraph}. Although they focus on learning correspondences between photos of objects in context situations and in iconic photos, they also discover a space that represents some notion of style. However, their notion of style is only based on visual similarity. Our work builds upon this approach, but extends it, because we want to learn a notion of style that goes beyond visual similarity. In particular, we want to learn the compatibility of bundles of items from different categories. Since this compatibility cannot be reduced to only visual similarity, we face a harder learning problem. To learn this compatibility we propose a novel strategic sampling approach for the training data, based on heterogeneous dyads of co-occurrences. To compare our framework to this approach, we include na\"ive sampling as a baseline in our evaluations. 
In particular, among the presented architectures in \cite{bell15siggraph}, we choose architecture B as baseline, because it gives the best results for cross-category search.

\textbf{Learning clothing style.}
There is a growing body of research that aims at learning a notion of style from images. For example, Murillo \etal \cite{murillo2012urban} consider photos of groups of people to learn which groups are more likely to socialize with one another. This implies learning a distance metric between images. However, they require manually specified styles, called `urban tribes'. Similarly, Bossard \etal \cite{bossard2013apparel}, who use a random forest approach to classify the style of clothing images, require pre-specified classes of style. In contrast, our learning framework learns a \emph{continuous} high dimensional space of style that does not require specified classes of styles. In a different approach, Vittayakorn \etal \cite{runway2realwayWACV15} learn outfit similarity, based on specific descriptors for color, texture and shape. While they are able to retrieve similar outfits to a query image, they don't learn compatibility between parts of outfits and, as opposed to our work, are not able to build outfits from compatible clothing items.

\begin{figure*}[t]
\begin{center}
   \includegraphics[width=0.75\linewidth]{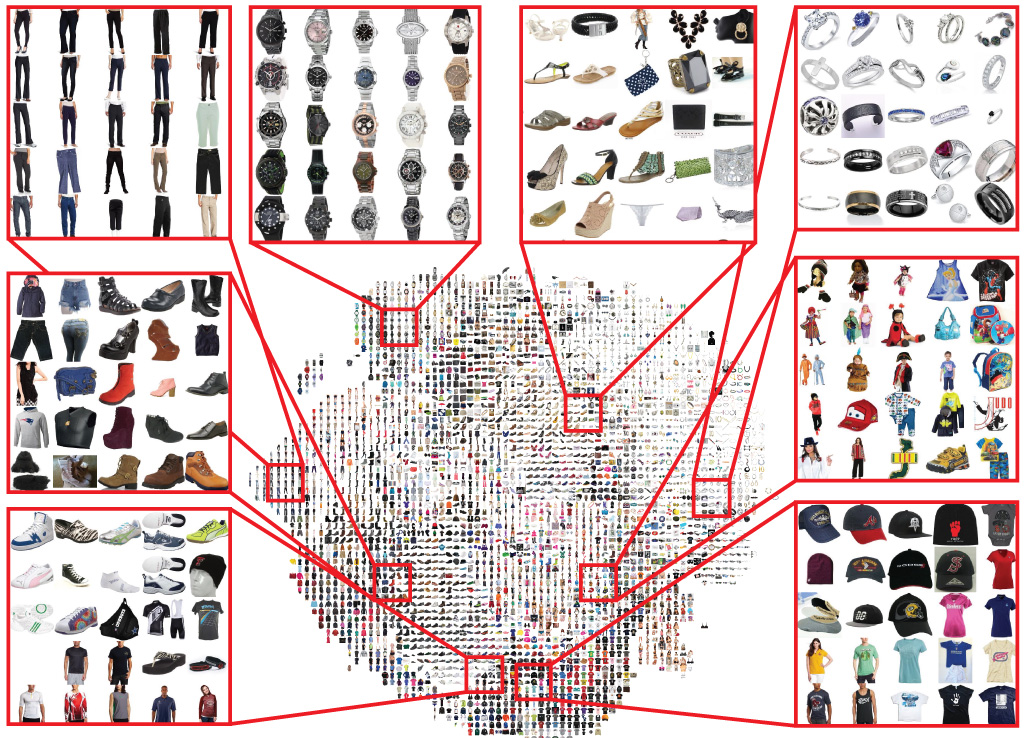}
\end{center}
\caption{Visualization of a 2D embedding of the style space trained with strategic sampling computed with t-SNE \cite{tsne}. The embedding is based on 200,000 images from the test set. For a clear visual representation we discretize the style space into a grid and pick one image from each grid cell at random. See the supplemental for the full version.}
\label{fig:embedding}
\end{figure*}

The closest work to ours in this line of research is the work by \cite{julian}. They collect the large scale co-purchase dataset from Amazon.com that we base our experiments on. Similar to our work, they also learn a notion of style and retrieve products from different categories that are supposed to be of similar style. However, their approach only uses the image features from the vanilla ImageNet AlexNet that was trained for object classification to learn their distance metric. Rather than using logistic regression, our approach goes further by fine-tuning the entire network with a Siamese architecture and novel sampling strategy.  Further, we demonstrate the transferrability of our features to an object category not seen during training.

\section{Dataset}
Training the Siamese CNN to learn the function $f$ requires positive and negative examples of clothing pairs. Let $t^{(+/-)}=(a,b)$ denote the training example containing items $a$ and $b$. Positive examples contain two compatible items $t^+=(a, b)$ s.t.~$\operatorname{comp}(a,b)$, whereas negative examples contain incompatible items $t^-=(a,b)$ s.t.~$\neg\operatorname{comp}(a,b)$.
Our learning framework requires items of positive training examples to belong to different categories, i.e., $t^+=(a,b)$ s.t. $\operatorname{comp}(a,b), a \in C_i, b \in C_j, i \neq j$.
However, publicly available datasets either don't contain labels for item categories, but only full body outfits \cite{runway2realwayWACV15} or do not provide information about item compatibility \cite{bossard2013apparel}.

Our work uses the large-scale dataset downloaded from Amazon.com by \cite{julian}. The dataset consists of three components: product images, product categories and product co-purchase information. While the dataset contains products from many categories, e.g.~books, electronics, clothing etc., we only consider the 'Clothing, Shoes, and Jewelry' category and its subcategories.
Following our notation, for each product $a$, the dataset contains one image $I_a$. Most of the images are iconic with a white background. However, some products are shown in full body pictures. Further, each product has a high-level category label $a \in C_i$, where $C_i \in \{\text{`pants'}, \text{`coats'}, ...\}$. We give a detailed overview of the distribution of the categories in the supplementary material. The advantage of using high-level categories such as `pants' or `shoes' is that they are independent of the notion of style, and thus, not subject to frequent change.

As a measure of compatibility between products, we use aggregated co-purchase data from Amazon. In particular, we define two items to be compatible, $\operatorname{comp}(a,b)$, if ``$a$ and $b$ are frequently bought together'' or ``customers who bought a also bought b''.
These are terms used by Amazon.com. Further, the relationships in the dataset do not come directly from the users, but reflect Amazon's recommendations \cite{linden2003amazon}, which are based on item-to-item collaborative filtering. 
For example, two items of similar style tend to be bought together or by the same customer. Many of the relationships in the co-purchase graph are not based on visual similarity, but on an implicit human judgment of compatibility. We expect the aggregated user behavior data to recover the compatibility relationships between products. However, there are challenges associated with using user behavior data, as it is very sparse and often noisy. While users tend to buy products they like, not buying a product does not automatically imply a user dislikes the item. Specifically in the Amazon dataset, two items that are not labeled as compatible are not necessarily incompatible.

\section{Learning the style space}
Given a query image, we want to answer questions like: ``What item is compatible with the query item, but belongs to a different category?''  More formally, let the query image be denoted by $I_q$ and the item depicted in the image be $q$. The membership of the item $q$ to a category $C_i$ is denoted by $q \in C_i$. Further, let $\operatorname{comp}(q,r)$ denote the boolean function that items $q$ and $r$ are compatible with one another. Then, our goal is to learn a function $r = \operatorname{retrieve}(I_q,j)$ to retrieve an item $r$ such that $\operatorname{comp}(q,r)$ and $q \in C_i, r \in C_j, i \neq j$. 
To retrieve compatible items, we learn a feature transformation $f: I_q \rightarrow s_q$ from the image space $I$ into the style space $S$, where compatible items are close together. Then, we can use the style space descriptor $s_q$ to look up compatible neighbors to $q$. 

The data on co-purchased items represents the aggregated preferences of the Amazon customers and defines a latent space that captures the customers' consensus on style. We are especially interested in the specific space that captures style compatibility of clothing items from different categories. Since Siamese CNNs learn a space defined by the training data, choosing the right sampling method of the training examples is important.

In this section, we first describe our novel sampling strategy to generate training sets that represent notions of style compatibility across categories. Then, we show how to train a Siamese CNN to learn a feature transformation from the image space into the latent style space. 

\subsection{Heterogeneous dyadic co-occurrences}
Two key concepts of the proposed sampling approach are \emph{heterogeneous dyads} and \emph{co-occurrences}. Generally, a dyad is something that consists of two elements, i.e., our training examples are dyads of images. Heterogeneous dyads are pairs where the two elements come from different categories. Formally, in the context of this work, a dyad is a pair of item images $(I_a,I_b)$ and a heterogeneous dyad is a pair $(I_a,I_b)$ s.t. $a \in C_i, b \in C_j, i \neq j$.

Co-occurrence generally refers to elements occurring together. For sales information, co-occurrence might refer to co-purchases, for food items it might mean that a group of items belong to the same menu or diet and for medical applications it might refer to symptoms often observed together. While this is a general concept, for our experiment, we define co-occurrence between items to be co-purchases.

\subsection{Generating the training set}
Before generating the training set, we remove duplicates and images without category labels. This reduces the number of images from $\approx 1.6$ million to $\approx 1.1$ million images. Training a Siamese CNN requires positive (similar style) as well as negative (dissimilar style) training examples.
To generate training pairs, we first split the images into training, validation and test sets according to the ratios $80 : 1 : 19$. When we split the sets, we ensure that they contain different clothing categories in equal proportions. Then, for each of the three sets we generate positive and negative examples. We sample negative pairs randomly among those not labeled compatible. We assume that these pairs will be incompatible with high probability, but also relatively easy to classify. We compensate this by sampling a larger proportion of negative pairs in the training set. In particular, for each positive example we sample 16 negative examples. 
Further, as pointed out by \cite{BellUpchurch15}, balancing the training set for categories can increase the mean class accuracy significantly. Thus, we ensure a balance of the positive examples over all clothing categories as much as size differences between categories allow. We choose a training set size of 2 million pairs, as it is sufficient for the network to converge. The validation and test set sizes are chosen proportionally.
We use three different sampling strategies:

\textbf{Na\"ive}: All positive and negative training examples are sampled randomly. Positive as well as negative pairs can contain two items from within the same category or from two different categories.

\textbf{Strategic}: 
The motivation for this sampling approach is the following: Items from the same category are generally visually very similar to each other and items from different categories tend to be visually dissimilar. For example all pants share many visual characteristics like their shape among each other, but are distinct from other categories like shoes. Further, convolutional neural networks tend to map visually similar items close in the output feature space. However, we want to learn a notion of style across categories, i.e., items from different categories that fit together should be close in the feature space. To discourage the tendency of mapping visually similar items from the same category close together, we enforce all positive (close) training pairs to be heterogeneous dyads. This helps pulling together items from different categories that are visually dissimilar, but match in style. Negative (distant) pairs can include both, two items from within the same category or from two different categories to help separate visually similar items from the same category that have different style.

\textbf{Holdout-categories}: The holdout training and test sets are generated to evaluate the transferability of the learned notion of style towards unseen categories. The training examples are sampled according to the same rules as in `strategic'. However, the training set does not contain any objects from the holdout-category. To evaluate the transferability of the learned style to the holdout-category, the test and validation set contain only pairs with at least one item from the holdout category.
\begin{figure}[t]
\begin{center}
   \includegraphics[width=1.0\linewidth]{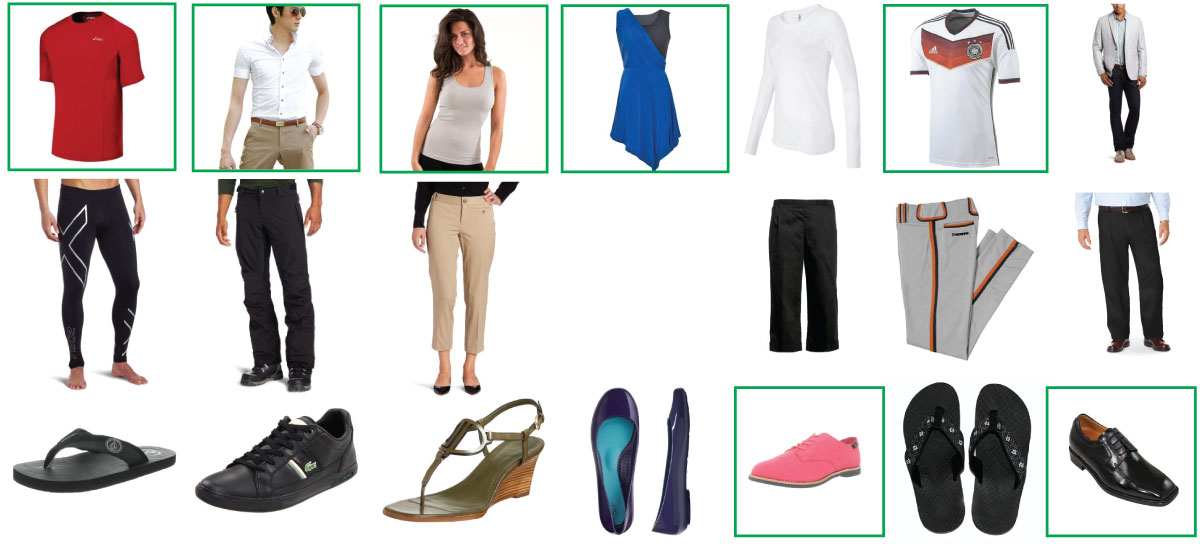}
\end{center}
\vspace{-6pt}
\caption{Each column: outfits generated with our algorithm by querying the learned style space. Query images are indicated by a green border. The other items are retrieved as nearest neighbors to the query item.}
\label{fig:outfits}
\end{figure}

\subsection{Training the Siamese network}
To train the Siamese networks, we follow the training procedure and network parameters outlined by Bell and Bala~\cite{bell15siggraph}. For more detailed background on training Siamese CNNs we refer to  Section~3 of~\cite{bell15siggraph}. As a basis for our training procedure, we use AlexNet and GoogLeNet, both pre-trained on ILSVRC2012~\cite{ilsvrc}, and augment the networks with a 256-dimensional fully connected layer. We chose 256, because \cite{bell15siggraph} show that 256 dimensions gave nearly the same performance as 1024 and 4096, but uses less memory. Then, we fine-tune the networks on about 2 million positive and negative examples in a ratio of 1 : 16. The training takes approximately 24 hours on an Amazon EC2 g2.2xlarge instance using the Caffe library~\cite{jia2014caffe}.

\begin{figure}[t]
\begin{center}
\begin{tabular}{@{}c@{}c@{}}
\includegraphics[width=0.5\linewidth]{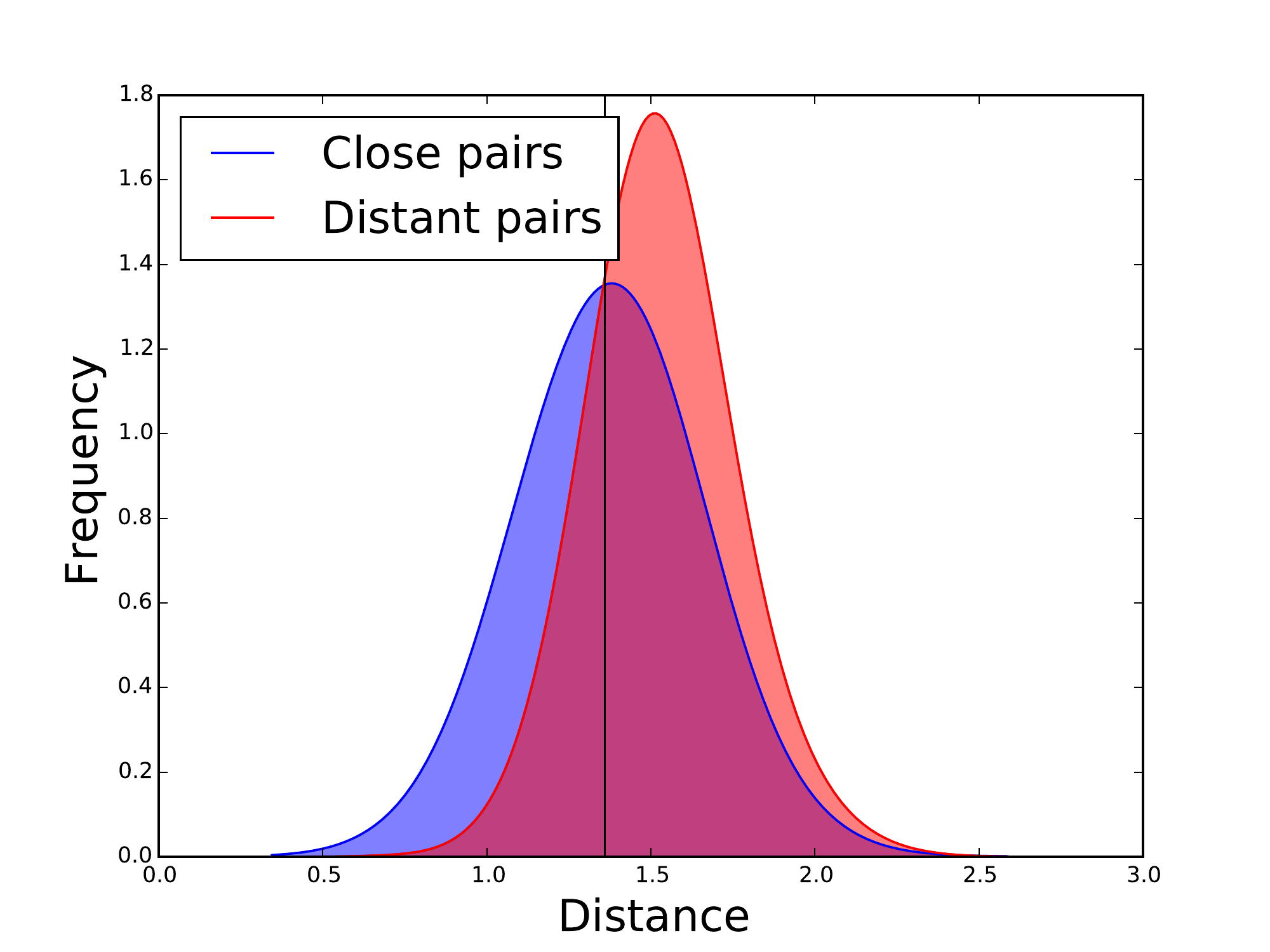} &
\includegraphics[width=0.5\linewidth]{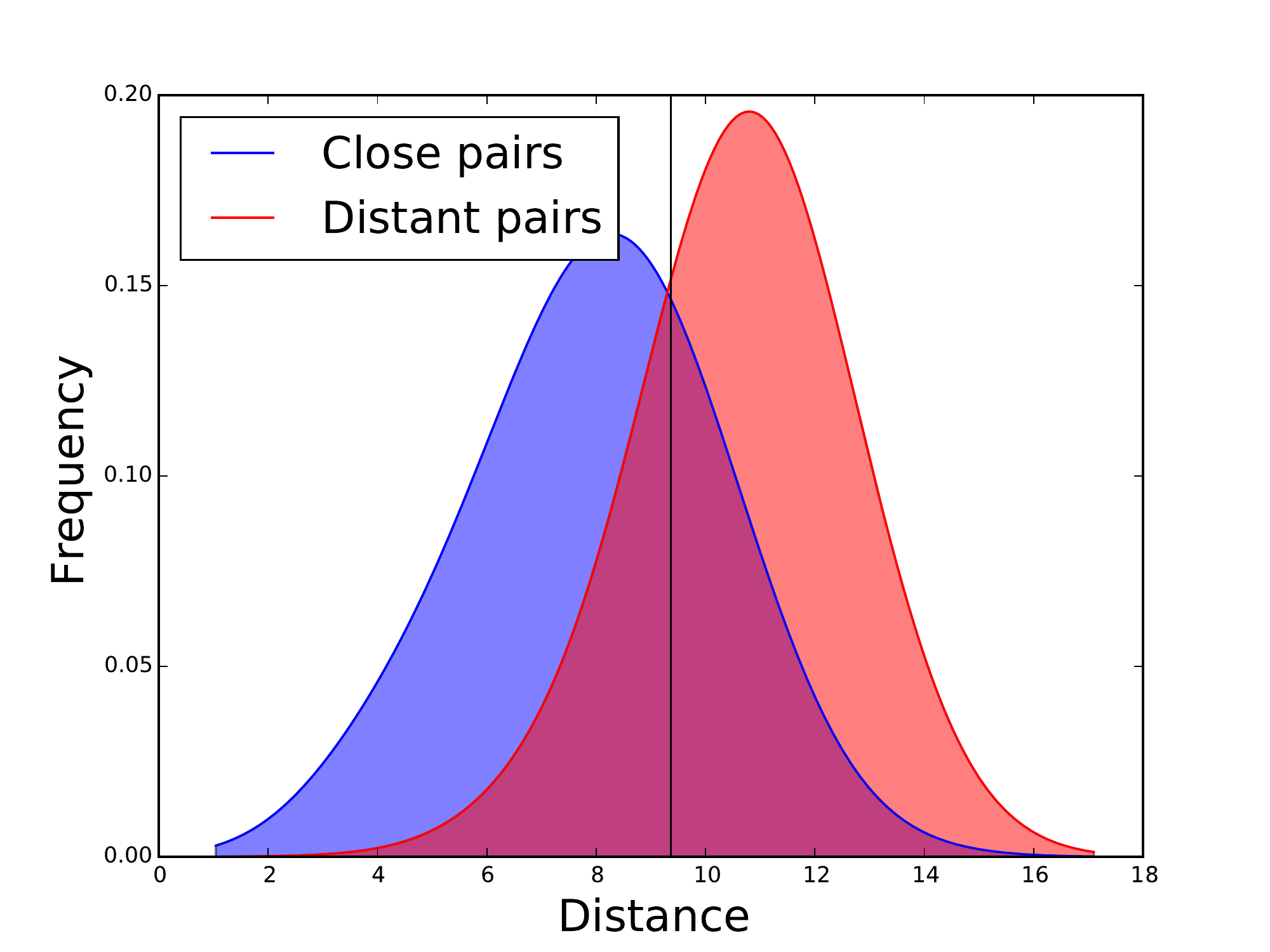} \\
(a) Before training &
(b) After training
\end{tabular}
\end{center}
\vspace{-14pt}
\caption{Distributions of distances for close and distant pairs for the vanilla GoogLeNet trained on ImageNet \textbf{(a)} and GoogLeNet trained with strategic sampling \textbf{(b)}. The fine-tuned GoogLeNet separates the two classes with a greater margin. This indicates that the network learned to separate matching from non-matching clothing.}
\label{fig:histograms}
\end{figure}

The objective of the network is to project positive pairs close together and negative pairs far apart. In Figure~\ref{fig:histograms} we plot the distributions of distances for positive and negative pairs for both before and after training. The plot shows that the fine-tuned GoogLeNet separates the two classes with a greater margin. This indicates that the network learned to separate matching from non-matching clothing.
\section{Generating outfits}
To generate outfits, we handpicked sets of categories that seemed meaningful to the authors. For example we did not sample ``dress'' and ``shirt'', because both cover the upper body and don't create a meaningful outfit.

Then, given the image $I_a$ of a query item $a$, we first use the trained network to project the image into the learned style space $s_a = f(I_a)$. For this explanation, let an outfit consist of the categories $C_{shoes}, C_{shirt}, C_{pants}$ and the query item is be a shirt, $a \in C_{shirt}$. Then, we look up the nearest neighbors to $s_a$ in space $S$ for the remaining categories $C_{shoes}$ and $C_{pants}$. One challenge with large scale datasets like the Amazon dataset is label noise, \eg 
, there exist shirts that are falsely labeled as shoes. Due to the tendency of Siamese CNNs to put similar looking objects close in the output space, this label noise gets particularly emphasized during nearest neighbor lookups. This means that with high probability a shirt labeled as shoe will be closer to the queried shirt than real shoes. To address this challenge we introduce a robust nearest neighbor lookup method. 
\begin{enumerate}
\item We use k-means to cluster the style space that contains all items labeled with the target category, in this example shoes. In our experiment, we use $k=20$ so that we get 20 centroids $\{c_1,\dots , c_{k}\}$. Then, we find the nearest centroid to the query item 
\begin{align}
c^* = \underset{c_i \in \{c_1,\dots., c_{k}\}}{\arg\min}\|s_a-c_i\|_2
\end{align}
\item Then, we sample a set of the $n$ closest items to the centroid. In our experiment, we choose $n = 5$. 
\begin{align}
\{x_1,\dots , x_n\} = \underset{s_i \in S, i \in C_{shoes}}{\arg\min(n)}\|c^*-s_i\|_2
\end{align}
\item Finally, we choose the closest item in $\{x_1,\dots , x_n\}$ to the query item for the outfit. 
\begin{align}
x^* = \underset{x_i \in \{x_1,\dots , x_n\}}{\arg\min}\|x_i-s_a\|_2
\end{align}
\end{enumerate}
This method allows us to robustly sample outfits and ignore images with false labels. Figure~\ref{fig:outfits} shows outfits generated by our algorithm.

\section{Visualizing the style space}
After we finished training the network, we can visualize the learned style space. First, we use the t-SNE algorithm \cite{tsne} to project the 256-dimensional embedding down to a two-dimensional embedding. We visualize the 2D embedding by discretizing the style space into a grid and picking one image from the images in each grid cell at random. Figure~\ref{fig:embedding} shows an embedding for all clothing categories. 

In this embedding, we can observe the notion of style learned by the network. Although the algorithm does not know the category labels, the embedding is mostly organized by object categories. This is expected behavior, as objects from the same category generally share many visual features. Further, the network is initialized with weights pre-trained on ImageNet, which separates the feature space by objects. However, we also observe that the network learned a notion of compatibility across categories. For example, the category of shoes has been split into three different areas dividing boots, flat shoes and high heels. 

In addition, we visualize stylistic insights the network learned about clothing that goes well together and clothing that does not. To do this, we first cluster the space for each category. Then, for each pair of categories, we retrieve the closest clusters in the style space (should be worn together) and the most distant clusters (ought not to be worn together).
Figure~\ref{fig:stylenotion} shows the close fits in the top rows and the distant clusters in the bottom rows. The clustering helps to avoid outliers that are incompatible to all types of clothing and further helps retrieve a general understanding of stylistic compatibility between groups of clothing. 


\begin{figure*}[t]
\begin{center}
\begin{tabular}{@{}c@{}c@{}c@{}c@{}}
\includegraphics[width=0.24\linewidth]{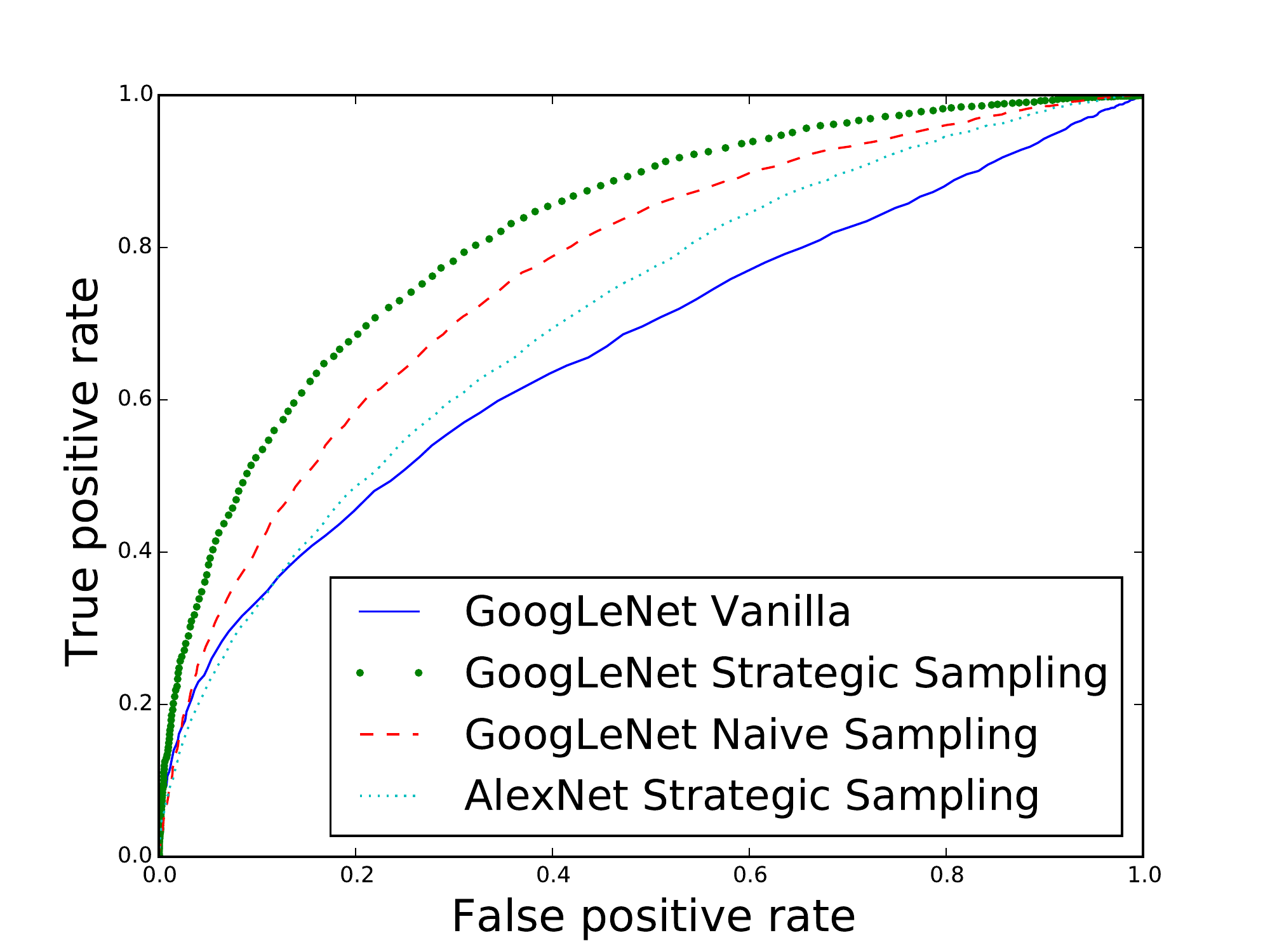} &
\includegraphics[width=0.24\linewidth]{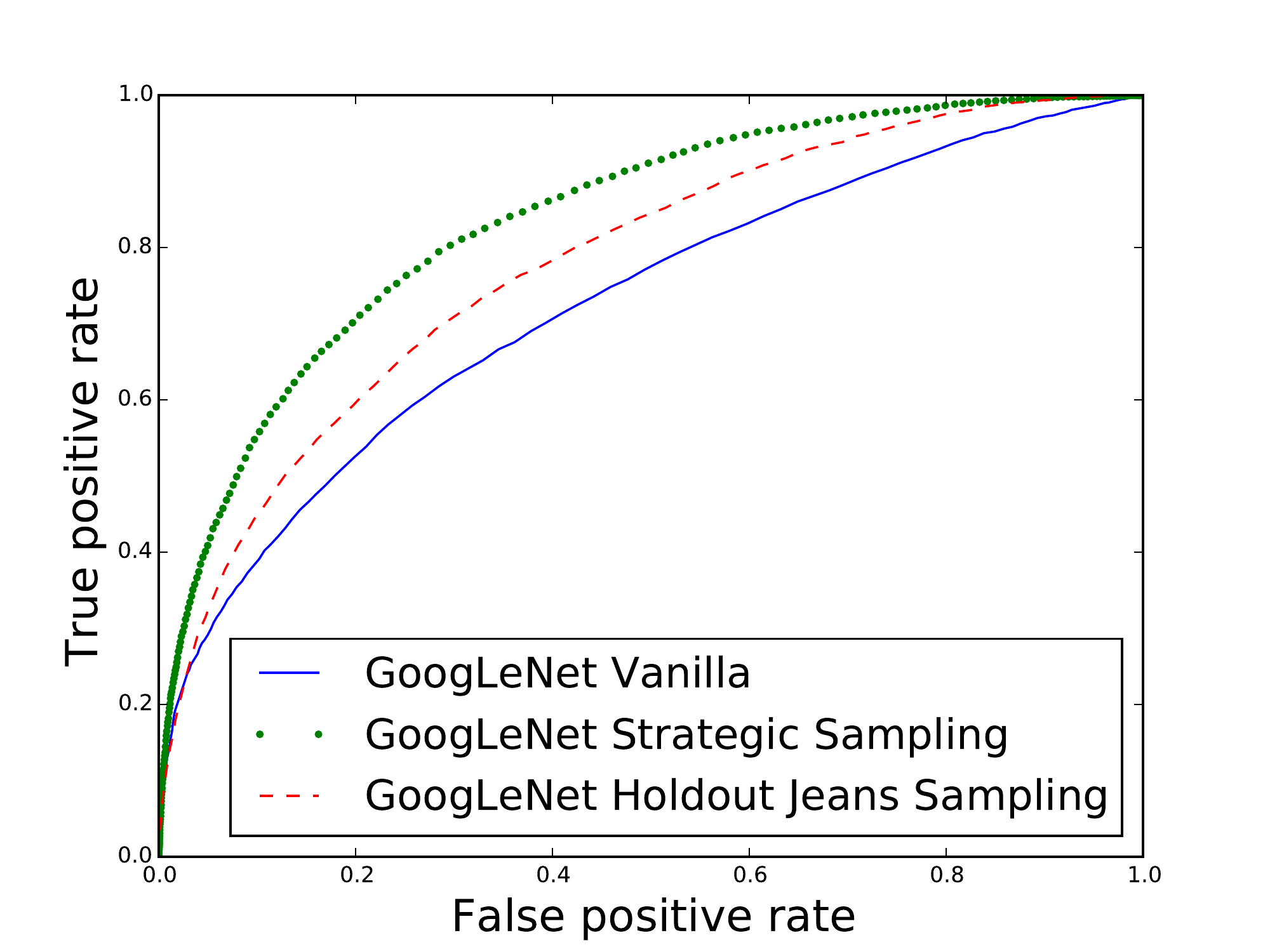} &
\includegraphics[width=0.24\linewidth]{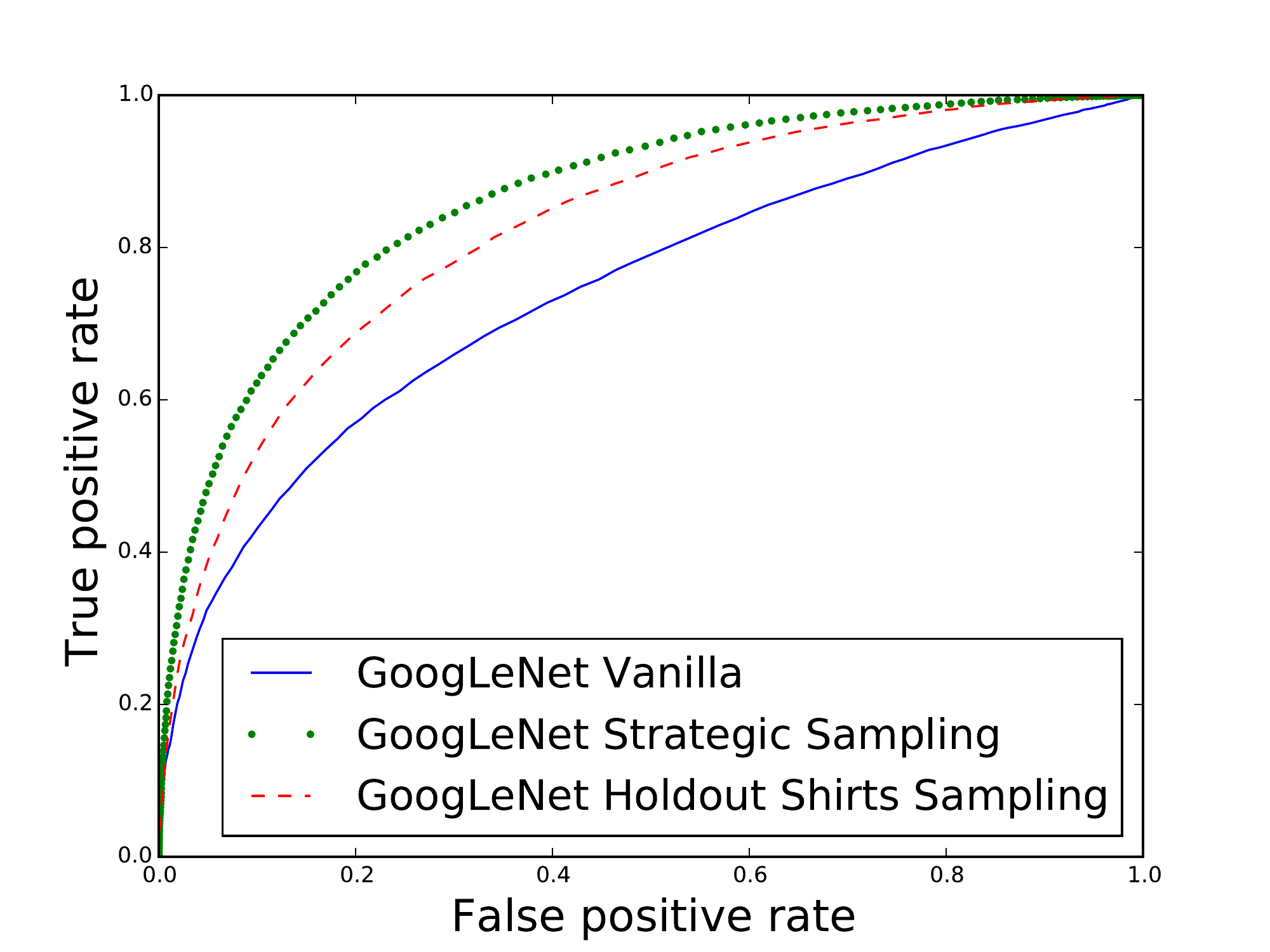} &
\includegraphics[width=0.24\linewidth]{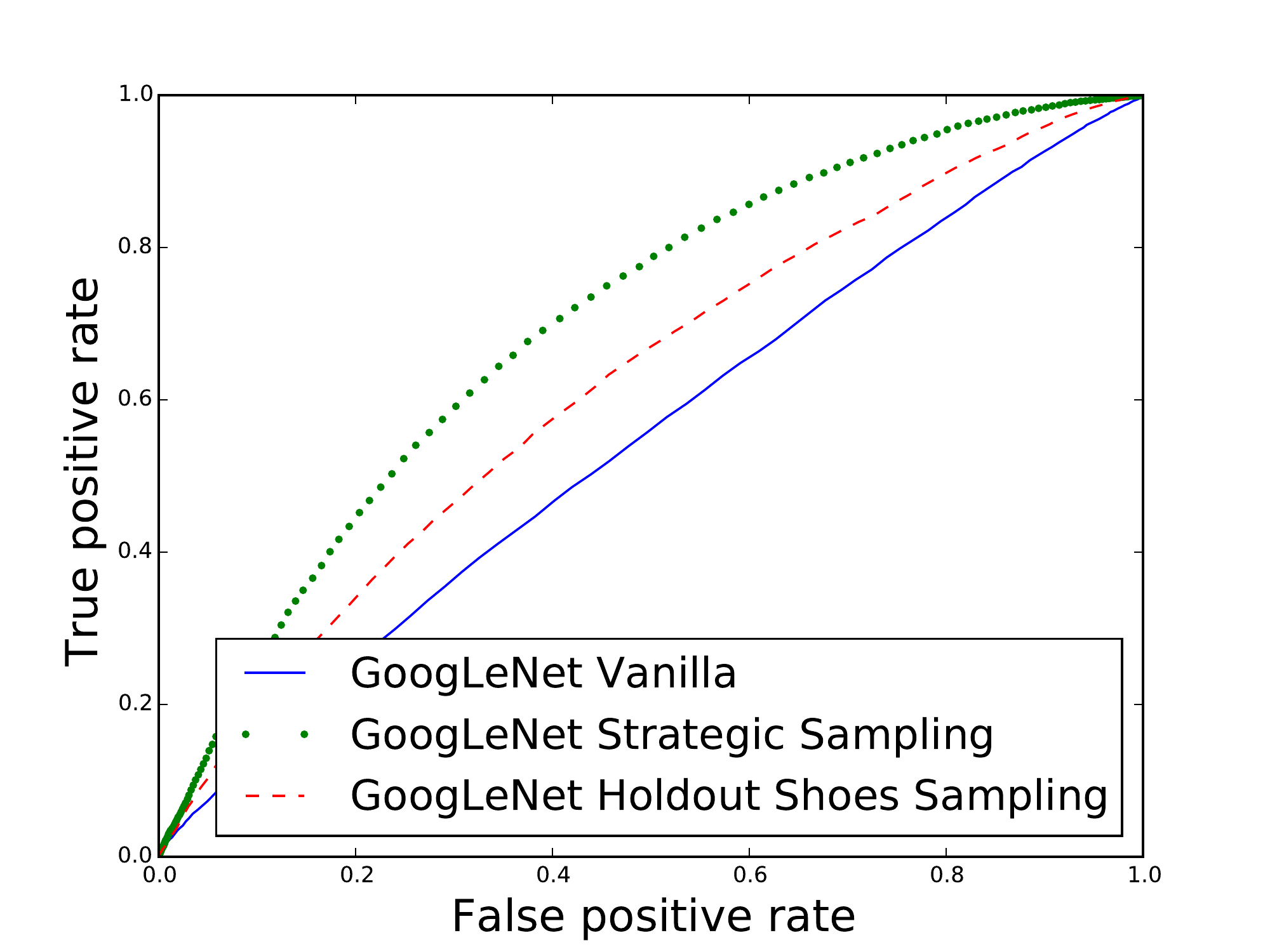} \\
(a) & (b) & (c) & (d)
\end{tabular}
\end{center}
\vspace{-12pt}
\caption{ROC curves of link prediction on the test set. (a) Test set and training set include pairs of items from all categories. Only `GoogLeNet Vanilla' was trained only on the ImageNet training set.
(b) Test set only includes dyads with at least a pair of jeans.
The training set for `GoogLeNet Holdout Jeans Sampling' does not include any jeans, whereas the training set for `GoogLeNet Strategic Sampling' includes items from all categories.
(c,d) Follow the same principle as (b), with holdout categories of shirts and shoes, respectively.}
\label{fig:roc-curves}
\end{figure*}

\section{Evaluation}
We perform four different evaluations for our proposed learning framework. First, we measure the link prediction performance on the test set and compare it against two baselines. Second, we evaluate the transferability of the learned features to new, unseen categories. Third, we compare against results in related work and lastly we conduct a large scale user study. 

\textbf{Test set prediction accuracy.} The first experiment measures the link prediction performance of our algorithm on the test set. The links in the test set are chosen according to the same rule as in the training set of our strategic sampling method. This means close links are all heterogeneous dyads whereas distant links can be both, within or across categories. Further the test set contains close and distant links in ratio $50:50$. We choose this test set as it measures the performance of predicting bundles of co-occurring items. We compare four different approaches: GoogLeNet and AlexNet, both trained with strategic sampling, GoogLeNet with na\"ive sampling as well as vanilla ImageNet-trained GoogLeNet. Figure~\ref{fig:roc-curves} shows the ROC curve computed by sweeping a threshold value to predict if a link is close or distant. The plot shows clearly that the strategic sampling approach outperforms all baselines. Table~\ref{tab:auc} shows the performance in terms of area under the curve (AUC). Since GoogLeNet outperforms AlexNet, we focus on GoogLeNet for the rest of the experiments.

\textbf{Feature transferability.} In the second setup, we evaluate the transferability of features learned with our framework to new unseen categories. To test transferability, we separate a holdout category from the training set so that the network has never seen items from the holdout category during training. Then, we test this network on a test set, where each link contains at least one item from the holdout category. We perform this experiment for three different holdout categories: shoes, jeans and shirts. The ROC curves for the different holdout experiments are shown in Figure~\ref{fig:roc-curves}. The results show that the networks that never saw items from the holdout categories clearly outperform the vanilla GoogLeNet baseline. In terms of AUC, the networks trained on the holdout testsets are able to achieve 47.5\% (shoes), 67.0\% (shirts) and 48.6\% (jeans) of the improvement of the network that saw the holdout category during training. This means that a great extent of the style features learned through our framework are transferable to unseen clothing categories. This is a very promising result as it indicates that our framework learns style features that generalize well to new categories.

\textbf{Comparison to related work.}
We also compare our method to \cite{julian}. Since the learning task as well as the training and test sets differ between their work and ours, the results are not directly comparable. In particular they learn and separately optimize two models, one to predict if items are \emph{bought together} and one to predict if they are \emph{also bought}. Further, their test sets contain mostly links within the same category: 85\% in `bought together' and 74\% in `also bought', whereas our model is specifically trained to predict links between categories. Despite the fact that we trained for a different task and did not distinguish between `bought together' and `also bought', we still get competitive results. 
In particular, we are able to achieve 87.4\% accuracy on bought together (compared to 92.5\%) and 83.1\% on also bought (compared to 88.7\%). 

\begin{table}
\begin{center}
\begin{tabular}{@{}l|c|c|c|c@{}}
	Methods         & All categ. & Shirts & Jeans  & Shoes  \\ \hline
	G-Net-vanilla   &  0.675  & 0.742 & 0.724 & 0.547 \\
	G-Net-na\"ive   &  0.770  & - & - & - \\
	A-Net-strategic &  0.721  &   -    &   -    &   -    \\
	G-Net-strategic &  \textbf{0.826}  & \textbf{0.865} & \textbf{0.836} & \textbf{0.700} \\
	G-Net-no-Shirts &    -     & 0.824 &   -    &   -    \\
    G-Net-no-Jeans  &    -     &   -    & 0.779 &   -    \\
	G-Net-no-Shoes  &    -     &   -    &   -    & 0.620
\end{tabular}
\end{center}
\vspace{-8pt}
\caption{AUC scores for all experiments. We can see that GoogLeNet with strategic sampling outperforms all other methods.}
\label{tab:auc}
\end{table}

\begin{figure}[t]
\begin{center}
   \includegraphics[width=1.0\linewidth]{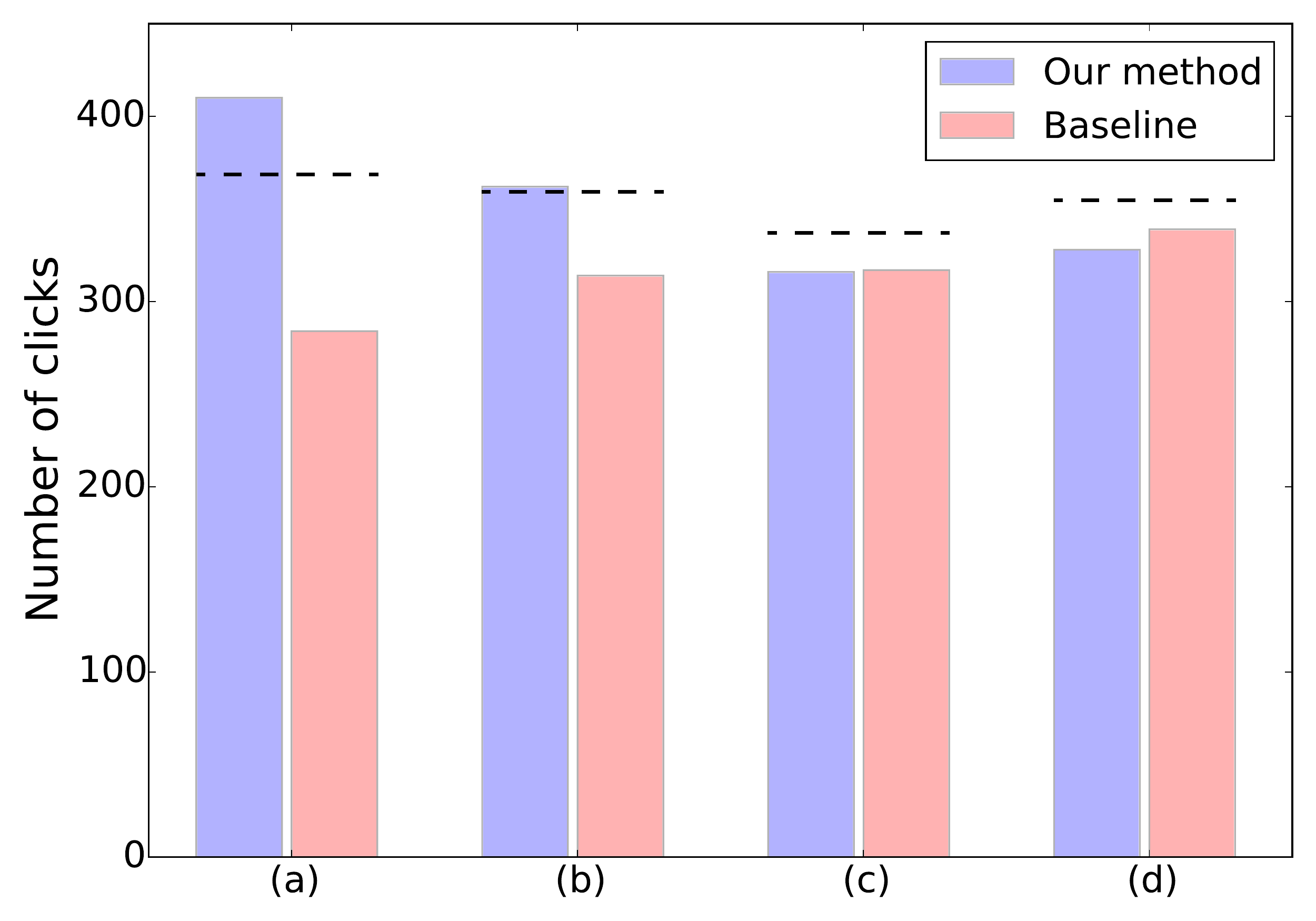}
\end{center}
\vspace{-10pt}
\caption{User study results. Our method GoogLeNet strategic compared against (a) random choice, (b) GoogLeNet na\"ive sampling, (c) AlexNet stragetic sampling and (d) GoogLeNet Vanilla.  Dashed line: if both bars are below this line, the difference is not statistically significant (hypothesis test for binomial distribution, 95\% confidence).}\label{fig:userstudy}
\end{figure}

\textbf{User study.}
Finally, we conduct an online user study to understand how users think about style and compatibility and compare our learning framework against baselines. During the study, we present three images to the users: One image from one category, \eg shoes, and two images from a different category, \eg shirts. Then, we ask triplet questions of the form:  ``Given this pair of shoes, which of the two presented shirts fits better?'' The two options for the user are predictions from different networks, using the previously described nearest neighbor retrieval method. We present these predictions in a random order, so the users cannot tell which algorithm's prediction they chose. The results of the study in absolute click counts are shown in Figure~\ref{fig:userstudy}.
We can see that for $(a)$ and $(b)$ the proposed method outperforms the baseline and the difference is statistically significant. For $(c)$ and $(d)$ we do not find a significant difference. Only studies $(a)$ and $(b)$ are consistent with the quantitative results. This indicates that the evaluations might measure different aspects of human behavior. While the quantitative results measure aggregated purchase behavior, the user study measures individual style preferences. 
We also conducted a survey with participating users asking them how they decide which option to pick.  The survey indicated that responses were not based only on stylistic compatibility; users were also influenced by factors such as:
\begin{enumerate}
\item Users tend to choose the option that fits in functionality, i.e. it serves the same function as the query item. For example, users pick long socks for boots, even though the colors or patterns don't match as well as the other pair of socks. 
\item Users sometimes choose the item that is stylistically similar, but not stylistically compatible.  For example, the query item and selected choice could have similar bright colors, but not actually match.
\item Users sometimes pick the item they like more, not the item that better matches according to style. 
\end{enumerate}

The survey indicates that, in addition to compatibility, users' decisions are also motivated by subjective preferences to individual items. This motivates future work in how to design user studies that can separate between hedonics and the perception of compatibility. 

\section{Conclusion}
In this work, we presented a new learning framework that can recover a style space for clothing items from co-occurrence information as well as category labels. This framework learns compatibility between items from different categories and thus extends the traditional approach of metric learning with Siamese networks that focus on recovering correspondences. Further, we present a robust nearest neighbor retrieval method for datasets with strong label noise. Combining our learning framework and nearest neighbor retrieval, we are able to generate outfits with items from different categories that go well together. Our evaluation shows that our method outperforms state-of-the-art approaches in predicting if two items from different categories fit together. Additionally, we conduct a user study to understand how users think about style and compatibility, and compare our learning framework against baselines. Furthermore we show that a great extent of the style features learned through our framework are transferable to unseen clothing categories. As future work, we plan to expand our approach to incorporate user preferences to provide personalized outfit recommendations.
\section{Acknowledgements}
\noindent
We would like to thank Xiying Wang for her assistance that greatly improved the illustrations.
We further thank Vlad Niculae, Michael Wilber and Sam Kwak for insightful feedback.
This work is partly funded by AOL-Program for Connected Experiences and a Google Focused Research award.

{\small
\bibliographystyle{ieee}
\bibliography{egbib}
}

\end{document}